# Exploring COVID-19 Related Stressors Using Topic Modeling

Yue Tong Leung, Farzad Khalvati

*Abstract*— **The COVID-19 pandemic has affected lives of people from different countries for almost two years. The changes on lifestyles due to the pandemic may cause psychosocial stressors for individuals, and have a potential to lead to mental health problems. To provide high quality mental health supports, healthcare organization need to identify the COVID-19 specific stressors, and notice the trends of prevalence of those stressors. This study aims to apply natural language processing (NLP) on social media data to identify the psychosocial stressors during COVID-19 pandemic, and to analyze the trend on prevalence of stressors at different stages of the pandemic. We obtained dataset of 9266 Reddit posts from subreddit \rCOVID19_support, from 14th Feb ,2020 to 19th July 2021. We used Latent Dirichlet Allocation (LDA) topic model and lexicon methods to identify the topics that were mentioned on the subreddit. Our result presented a dashboard to visualize the trend of prevalence of topics about covid-19 related stressors being discussed on social media platform. The result could provide insights about the prevalence of pandemic related stressors during different stages of COVID-19. The NLP techniques leveraged in this study could also be applied to analyze event specific stressors in the future.**

## I. INTRODUCTION

T HE COVID-19 pandemic has affected lives of people from different countries for almost two years. To prevent the risk of infection, countries have implemented different safety measures such as social distancing, lockdown, and school closure. Individuals' mental health has been impacted due to experiencing stress, anxiety, loneliness, and feeling uncertain about the pandemic. Recognizing the common psychosocial stressors and their prevalence at different stages of the pandemic will allow the healthcare providers and social workers to provide high quality mental health interventions and support. Several studies have been done to understand the impacts of COVID-19 on mental health [1][2][3]. The majority of the studies used questionnaires or interviews to obtain data for analyzing the mental health impacts of the pandemic. However, the process of obtaining data could be time-consuming and costly. As the development of the pandemic is evolving rapidly, individual's mental health status could also be changing quickly. The survey methods may not be able to capture the latest needs of mental health support. In contrast, social media may allow close to real-time monitoring of the mental health impacts during the pandemic, as individuals actively share their feelings and difficulties on social media platforms. In this study, Natural Language Processing (NLP) techniques are applied to identify COVID-19 related psychosocial stressors that were discussed on social media, and to visualize the prevalence of stressors at different stages of the pandemic.

Social media platforms, such as Twitter and Reddit, are commonly used as the data source for obtaining insights regarding the mental health status. As people will share their feelings or experiences on the platforms, the contents on social media may reflect users' emotions. The changes on emotions of the population could be reflected on their behavior on social media. Many researchers have utilized NLP techniques and social media data to analyze mental health status for the population. De Choudhury et al. [4] have introduced 'social media depression index (SMDI)' to monitor the trends of depression across the population. Larsen et al. [5] have devised a system 'We Feel' which applies sentiment analysis and data visualization to tweets to obtain real-time insights of emotional states for the population. Compared to the traditional methods such as surveys and interviews, social media can provide close to real-time data for analyzing the latest status of mental health for the population. While many research devises methods to utilize social media data to measure population mental health status, few of them focus on identifying the psychosocial stressors which are mentioned on social media. According to American Psychological Association [6], a psychosocial stressor is defined as "a life situation that creates an unusual or intense level of stress that may contribute to the development or aggravation of mental disorder, illness, or maladaptive behavior". Mowery et al. [7] have developed an annotation scheme to identify depression symptoms and psycho-social stressors, such as problems with expected life course and problems with primary support group, mentioned in tweets. Feldhege et al. [8] have used topic modelling to identify the prevailing topics, such as 'motivation', 'ending relationship', 'studying/university' and 'therapy', that were mentioned on mental health support groups of Reddit. If we could identify

Y. Leung is with Department of Mechanical and Industrial Engineering, University of Toronto (email: yuetong.leung@mail.utoronto.ca)

F. Khalvati is with Department of Mechanical and Industrial Engineering, Institute of Medical Science, and Department of Medical Imaging, University of Toronto, and Department of Diagnostic Imaging, Neurosciences & Mental Health Research Program, The Hospital for Sick Children, Toronto, Canada (email: farzad.khalvati@utoronto.ca)

stressors mentioned in the posts, we could have a better understanding on the changes in the prevalence of stressors at different stages of the pandemic. In this study, Reddit is selected to be the data source for applying machine learning model to obtain insights about stressors during COVID-19. There are several advantages of using Reddit in this study. De Choudhury and De [9] found that the anonymity of throwaway accounts on forum like Reddit can promote self-disclosure about mental health issues. Naseem et al. [10] found that COVID-19 related discussions on Twitters focused on news and opinions about government policies, while discussions on Reddit focused on how pandemic has affected people's lives. As a result of these tendency, Reddit has high potential to be a better data source for identifying psychosocial stressors during the pandemic.

In this study, we use Latent Dirichlet Allocation (LDA) topic model for identifying the COVID-19 related distress that were mentioned on Reddit. Topic model is an unsupervised machine learning model which can be applied to different research topics such as computational social science and understanding scientific publications [11]. LDA is a topic model that is commonly used to summarize the topics on social media. LDA model is based on the following assumptions: first, a topic is a combination of terms with a probability distribution; second, a document is generated by a combination of topics with a probability distribution [11]. The algorithm starts by setting topic assignments randomly; then computes the distribution of words in a topic and the distribution of topics in a document. Then, update the topic allocation of words and iterate until convergence. The LDA model will return the distribution of topics in each of the documents and the distribution of words in each of the topics. With the topic allocation of documents, the dominant topic of each of the posts in our dataset could be determined.

In the field of public health and informatics, some researchers utilized topic modelling to summarize the COVID-19 related discussions on social media. Medford et al. [12] have observed the trend of topics in COVID-19 related tweets in the early stage of the outbreak. Jelodar et al. [13] have leveraged LDA to identify the topics that are being discussed on COVID-19 related subreddits such as \rCOVID19, \rCoronavirus and r/CoronaVirus2019nCoV. Jang et al. [14] have used LDA to identify the topics on Twitters to understand the reactions and concerns about COVID-19 in North America. Besides summarizing the pandemic related discussions, some research has leveraged NLP techniques on social media data to analyze the mental health impacts of COVID-19. Biester et al. [15] have observed the changes in Reddit mental health support groups after the outbreak of COVID, and obtain insights about the impacts on individuals' mental health. In that study, LDA topic model was applied to identify topics such as family and school in the subreddits, and then leveraged time series analysis to find out which of the topics were affected after the outbreak of the pandemic. Low et al. [16] have used different NLP techniques to characterize the differences on Reddit mental health support groups in pre-pandemic period and mid-pandemic period.

In this paper, we utilize LDA topic model to identify the pandemic related distress, by identifying the topics being discussed on subreddit \rCOVID19_support. After applying LDA model, we have visualized the trends on prevalence of topics at different stages of the pandemic. Users on the platform \rCOVID19_support could ask questions about COVID-19, and share the experience during the hard time in the pandemic. Due to the topics focused in this subreddit, we consider this platform to be a rich source for understanding the difficulties faced by and the feelings of the population during the pandemic. Several existing research has leveraged NLP to analyze the population mental health status during the pandemic; however, the existing studies has not explored the possibility of analyzing COVID-19 related stressors. This study focuses on monitoring the stressors during the pandemic. While most of the existing research focuses on the mental health impacts in the beginning of outbreak of the pandemic, the dataset extracted in this study covered the posts on the subreddit starting from the outbreak of the pandemic up to July 2021. This allows us to visualize the changes on prevalence of stressors at different stages of the pandemic. Observing the trends can provide insights of what are the latest predominant stressors, and the findings could also be useful for mental health support providers and policy makers. We believe applying NLP to summarize texts can help us obtain insights about the mental health status and stressors during the pandemic.

## II. Methods

### A. Data Collection and Pre-Processing

Reddit posts were extracted from subreddit \rCOVID19_support using Pushshift API [17]. The dataset included 9266 posts from 14 Feb 2020 to 19 July 2021. The posts which were marked as '[removed]' or '[deleted]' were excluded from the dataset. On Reddit, some of the posts are tagged by "flairs", which describe the contents or nature of the posts. The flairs in the dataset obtained includes: "Support", "Questions", "Discussions", "Trigger Warning", "Good News", "Firsthand Account", "Resources", "Vaccines are SAFE", "News", "Biosafety Request", "The answer is NO", "Misinformation-debunked", and "Desperate mod". Posts that are tagged with flairs such as "Resources" and "News" have low tendency to include contents related to psychosocial stressors. With the use of flairs tagged to the posts, we could filter the posts that have low tendency of including contents that are related to our research topics. On the subreddit, some of the posts included contents of sharing information such as latest news about COVID19, potential adverse effects of vaccines and tips about infectious prevention. Some users asked questions such as which vaccines are safe, whether the adverse effects of vaccines are normal, and whether it is safe to visit grandparents during the pandemic. The posts which labelled with flairs 'Support' and 'Trigger Warning' have high tendency to include contents about users' personal experience, stressors or feelings during the pandemic. Posts with flairs "News" and "Questions" tend to not include contents about stressors. As we focus on understanding stressors in this study, we have extracted a subset to include posts which labelled with flairs that have high tendency to include contents about stressors.

In the dataset, 4654 posts were labelled by flairs and 4612 were not. The number of posts tagged by each of the flairs are shown in Table 1. Missing flairs were predicted using a logistic regression model that was trained by the labelled data. Before training the classifier, the texts in the posts were represented by term frequency-inverse document frequency (TF-IDF), which quantifies the significance of a given term compared to the other terms within the given document and within the given corpus [18]. The features used to train the classifier included LDA features (n=10), TF-IDF features (n=200) and one feature to describe whether the posts include hyperlink (n=1). After filling the missing flair in dataset, 5472 data points were labelled by flairs that are likely describing stressors during COVID-19. These data points were extracted as subset for analysis.

TABLE 1
FLAIRS TAGGED TO THE POSTS IN THE DATASET

| Flair Groups | Flair | Number of posts | | |
|---|---|---|---|---|
| | | Subset with labelled flair (n=4654) | Subset with predicted flairs (n=4612) | Dataset with labelled or predicted flairs (n=9266) |
| Mental health support | Support | 2386 | | |
| | Trigger Warning | 197 | | |
| | | (2584) | 2888 | 5472 |
| Discussion and Questions | Questions | 1069 | | |
| | Discussion | 597 | | |
| | Vaccines are safe | 55 | | |
| | | (1742) | 1417 | 3159 |
| News and Resources | Good News | 146 | | |
| | Resources | 59 | | |
| | News | 18 | | |
| | | (226) | 225 | 451 |
| Experience | Firsthand account | 102 | (102) | 82 | 184 |
| Others | Biosafety request | 14 | | |
| | The answer is NO | 7 | | |
| | Misinformation debunked | 3 | | |
| | Deperate mod | 1 | | |
| | | Posts removed as only few posts were labelled with these flairs | | |

### B. Topic Model Training and Evaluation

In this study, LDA model in scikit-learn package was used to identify the topics in dataset. The model was trained by TF-IDF features which were created from the dataset. For each of the data points, topic model has output the proportions of contents belonging to each of the topics. The dominant topic for each of the posts was then identified by finding the topic with highest value in LDA output. For each of the LDA topics, 3 sample posts with the highest LDA output percentage and 3 random posts with corresponding dominant topics were selected for manual review. Topic coherence was manually evaluated by reviewing the data points selected. If the sample data points of same dominant topics had similar contents, the topic could be named. After naming the LDA topics, some topics shared similar contents with each other. LDA topics were then grouped into "topic groups". Six topic groups have been identified: educational and occupational problems, family problems, fear of coronavirus, mental health symptoms, problems related to social environment, and uncertainty on development of pandemic.

### C. Feature Engineering

The output of topic modelling is highly dependent on the feature vector (a matrix of TF-IDF values for each of the documents). For the first trial, the feature vector used to train LDA model included TF-IDF with max_feature 300, which means the feature vector included 300 columns of tokens, which are the terms (can consist of one or more words) that has highest TF-IDF values in the given corpus. Then, the model was evaluated by the methods mentioned before, which is selecting sample posts from each of the topics and evaluate the topic coherence manually. Besides evaluating LDA topic coherence, features were manually evaluated to determine whether they were likely to cause LDA topic model to cluster posts in desired ways. For example, topic model may group posts with tokens 'suggestion', 'anyone' and 'thank you' into the same topic because those words tend to appear together when authors were asking for suggestions at the end of posts. However, clustering this topic do not help in understanding the stressors, and may act as noise. To avoid this noise, those tokens were removed from the feature vectors. Beside removing the tokens, some words may be useful for identifying topics. For examples, the tokens 'grocery shopping', 'maskless' and 'no mask' were commonly appeared when users were expressing their fears of getting infected. Including those words into the feature vector could help the LDA topic model to identify topics related to fear of coronavirus. The feature vector was then updated by selecting the tokens. Then the LDA was trained by the new feature vector, leading the output of model to be closer to the desired result. The iteration process of improving topic model is illustrated in **Fig. 1**. The iteration was ended when the sample posts shared similar contents within the same LDA topic. By this iteration of feature selection and evaluation of topic model output, the performance of topic model was controlled.

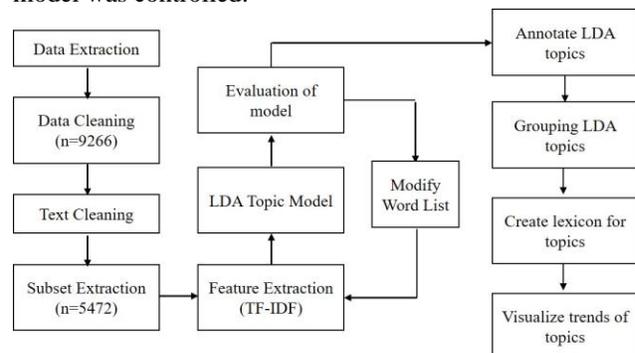

Fig. 1 Overview of research framework

### D. Psychosocial Stressor Lexicon

After the topic groups were defined, each of the topic groups was evaluated for searching keywords that can directly indicate the existence of topics in the text. For example, if the posts include education related words such as "college" and "online learning", it can be concluded that the authors have mentioned educational problems in the post. Lexicons were created for some of the topics. If a post includes any words listed in the lexicons, it was assumed that the post included contents related to corresponding topics. Each of the posts could contain more than one topic. With the use of lexicons, each of the posts was annotated by whether it includes contents of each of the topics.

In LDA, topics are defined as a mixture of terms with different probability distribution; this means a word could belong to more than one topic and it could cause inaccuracy on the prediction. In contrast, lexical approach has higher interpretability on topic classification; but it requires careful selection of keywords to avoid the terms that could belong to more than one topic. In this study, we assumed we do not have prior knowledge on how Reddit users expressed their feelings on the subreddit. To obtain insights about what topics existed in the subreddit and what could be the keywords for each of the topics, we applied LDA model before applying the lexical approach. In this study, lexical approach was created for two purposes. First, to further analyze the subtopics within topic group. Some of the topics may include some common words, and have to be grouped into the same topic group in LDA model. The topics in the same topic groups could be separated by choosing unique keywords. The second reason for creating lexicon is to verify the result from LDA.

### E. Visualization

In previous parts, each of the data points was annotated with the LDA output, which represents the proportion of contents belonging to each of the topic groups, and the output of lexicons, which represents whether the post includes words that were listed in lexicons. Then the monthly sum of LDA model output for each topic groups were computed. The trend of topics in dataset can then be visualized and compared with the development of pandemic. Regarding the pandemic development, the numbers of total cases, new cases per day and the vaccinated population were obtained from Our World Data [19]. In this study, only numbers of The United States, United Kingdom and Canada are included because the majority of Reddit users are from these countries [20].

### III. RESULTS

#### A. Topics identified by topic model

After grouping LDA topics, six topics were identified: "fear of coronavirus", "educational and occupational problems", "family problems", "problems related to social environment", "mental health symptoms" and "uncertainty on development of pandemic". Fig. 2 shows the word clouds for each of the topics.

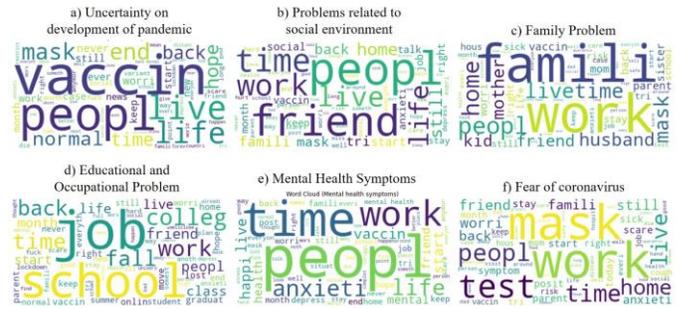

Fig. 2 Word-Cloud for each of the topics identified: a) Uncertainty on the development of the pandemic, b) Problems related to the social environment, c) Family Problem, d) Educational and Occupational Problem, e) Mental Health symptoms, f) Fear of the corona

The major topic in the dataset was "fear of coronavirus". Taylor et al. [3] has devised "COVID Stress Scale" to measure COVID-19 related distress and identify people in need of mental health services. The scale includes symptoms such as "danger and contamination fears" and "compulsive checking and reassurance seeking". The study has found that some people may have worries such as "social distancing is not enough to keep me safe from the virus", "people around me will infect me with the virus" and "can't keep my family safe from the virus". In our dataset, those worries were also expressed, and were identified by the LDA topic model. The words frequently appeared in the topic of "fear of coronavirus" is shown in **Error! Reference source not found.**. The words "home", "live" and "family" were frequently appeared when the authors of posts explained they were living with family members, followed by expressing the worries of infecting or getting infected by the family members. The words "mask", "risk" and "scare" were appeared when the authors shared the worries of getting infected on street.

LDA model identified topic of "educational and occupational problems". Some students had shared their feelings about online learning experience and some felt upset because of missing social engagement on school activities. Some people expressed the worries of finding jobs during pandemic. Some of the posts mentioned the situation of current university student having online classes for the final year and worried about finding a job after graduation. However, due to existence of this type of contents, tokens "online learning", "graduation", "university" and "find jobs" appeared together in the posts, and we grouped the educational and occupational LDA topics into the same group.

The topic "family problems" included contents about several subtopics: worry of elderly parents infected; worry of parents who already infected; anger of having opposing views with family members due to different acceptance on social distancing measures. The topic "mental health symptoms" included mentioning symptoms such as insomnia, Obsessive Compulsive Disorder (OCD) and feeling depressed. Some of the users described the feelings and mental health symptoms on Reddit and asked for mental health support, but did not explain the stressors. The topic "problems related to social environment" included contents about loneliness and uncertainty about balance between social activities and pandemic safety measures. The topic "uncertainty on development of pandemic" refers to the contents about the worry of the pandemic will last forever. For this topic, some posts included discussions about whether vaccination can help getting back to normal life.

### B. Trends of LDA topic groups

On 11 Mar 2020, World Health Organization has declared COVID-19 as "pandemic" [21]. The LDA output of the topics has been shown in Fig. 3. In Mar 2020, the number of posts on the subreddit was the highest. Then, the number declined from Apr 2020 to Jun 2020. In Sept 2020 to Dec 2020, the number rose steadily. The number of posts related to stressors was then declined from Jan 2021 to Mar 2021, while vaccinated population was growing.

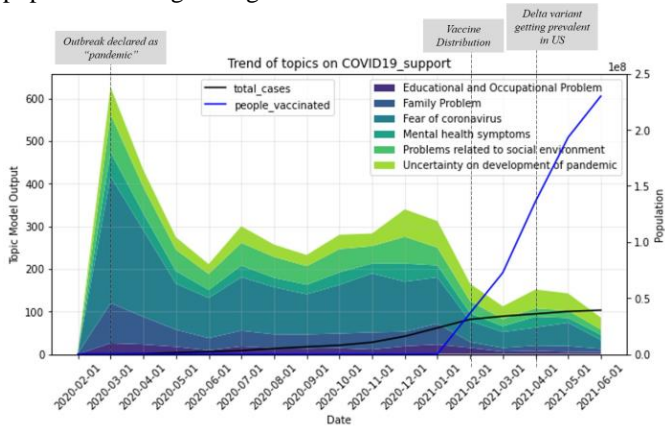

Fig. 3 Trend of topics on \rCOVID19_support. Stacked Area plot represents the sum of LDA output for each month, for each of the topics. Line plot represents the total number of cases and vaccinated population.

As shown in Fig. 4, the number of posts mentioning fear of coronavirus did not have significant changes from May 2020 to Dec 2020, although the daily number of cases had a significant increase in Nov2020. This suggests the fear of coronavirus was independent to the actual number of cases. The prevalence of discussion on fear of coronavirus was the highest in Mar 2020, when COVID-19 was first declared as 'pandemic'. After Feb 2021, the discussions about fear of coronavirus reduced. This suggests the distribution of vaccines reduced the perceived risks of infection of the coronavirus. Except "uncertainty on development of pandemic", all of the topics had a drop in prevalence after the vaccine was distributed in Feb 2021. For the topic about uncertainty, its prevalence in mid-vaccine stage is similar to pre-vaccine stage. The prevalence of uncertainty was the highest in Dec 2020 and Jan 2021.

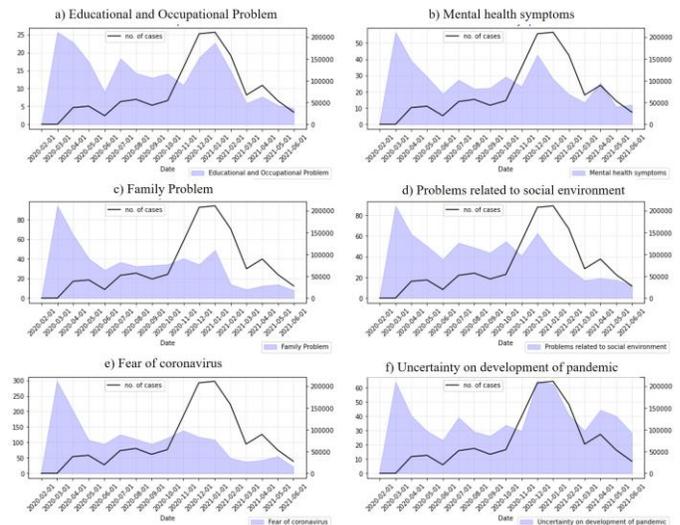

Fig. 4 Trend on prevalence of topic groups. Line plot represents the number of cases for each month. Area plot represents the prevalent of topics, which was measured by output of LDA.

Fig. 5 displays the proportion of each topic with respect to all topics. As it is shown, "Fear of coronavirus" was the major discussions on the platform. From Mar 2020 to Nov 2020, more than 40% of contents on the forum belonged to the fear of coronavirus. Starting from Dec 2020, the proportion was below 40% and showed a decreasing trend, while the proportion of posts about "Uncertainty on development of pandemic" steadily increased. On Jun 2020, proportion of the topics about pandemic development was higher than the topics about fear. For other topics, the prevalence did not have significant changes. The result suggested that the major stressors about COVID-19 has shifted from fear of getting infected into feeling uncertain about the development of pandemic.

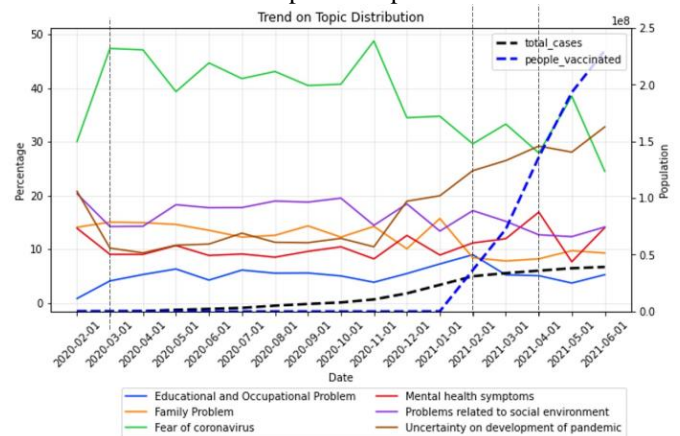

Fig. 5 Trend on proportions of topics mentioned on \rCOVID19_support. Dashed line represents the total number of cases and vaccinated population. Solid line represents the proportion of each topic for each month.

### C. Lexicon Approach

Table 2 shows the lexicon created based on the result of LDA model. The lexicons were created for two purposes. First, to further analyze the subtopics within topic group. As explained in previous sections, the LDA topics of occupational problems and educational problems had to be grouped together. By reviewing the word-clouds and sample texts of the topic group, we listed the words that could identify the educational and occupational problems separately. Therefore, we could analyze the prevalence of each of these subtopics. The second reason for creating lexicon was to verify the result from LDA. Lexicons about coronavirus and pandemic development have been created for verifying results from LDA. For example, according to randomly selected sample posts in each topic group, posts which include tokens "no mask" and "without mask" have mentioned the worries of seeing people do not wear masks on street and the authors worry about getting infected. Since the topic annotations with lexicon approach are more explainable, it was used for verifying LDA. After annotating the topics, the trend on prevalence of topics with LDA approach and lexicon approach had been visualized for comparison. According to the result as shown in Fig. 6, the result using the two approaches were comparable. The Person correlation coefficient of result for topic "Fear of coronavirus" of two approaches is 0.995; the correlation coefficient of "Pandemic Development" is 0.829.

The keywords for each of the topics were selected by evaluating the sample posts from the topic groups in LDA model. As we assumed we have no prior knowledge on what mental health issues were expressed on the platform and what keywords were commonly used for each of the topics, LDA model was applied before using the lexical approach. Once obtained the keywords and topics, we could use lexical approach to label the topics mentioned in each of the posts, and visualize the trend. In some cases, the posts may describe the topics without using the keywords. For examples, the topic "mental health symptoms" in LDA model, the texts in a post may include the words "feel", "anxious", "depressed", "tired"; however, those words are also common in other topics and may not be suitable to include those words in lexicons. For this case, LDA model is needed to identify that topic. In this study, both methods are used, and have shown the results for topics "Fear of coronavirus" and "Pandemic Development" are similar (measured by correlation coefficient).

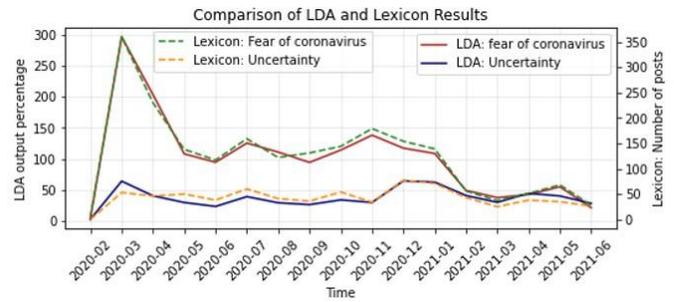

Fig. 6 This figure compares the trend on prevalence of topics. Solid line refers to the result obtained by LDA model, and dashed line refers to the result obtained using lexicon

Fig. 7**Error! Reference source not found.** shows the result of lexical approach for each of the topics. According to Fig. 7**Error! Reference source not found.**, the number of posts mentioning education problems had a declining trend from Mar 2020 to Jun 2020. In Mar 2020, the peak could be due to the start of school closure, when both students and teachers were not familiar with online learning. After that, the students may be adapted to the situation. The prevalence of stressors regarding online learning may be declined when the date was closer to summer holiday. However, the prevalence of the discussions was then increased from Jun 2020 to Oct 2020. This may suggest that prevalence of stress increased with time. Then, the prevalence was decreasing, starting from Oct 2020. Compared to other topics, the prevalence of education related discussions started decreasing before the vaccine distribution. The trend was independent of the number of cases. According to Fig. 8, in Sept 2020, more than 40% of posts mentioned about loneliness. After Sept 2020, there was a decreasing trend on the topics about loneliness.

TABLE 2
COVID-19 STRESSORS LEXICON

| Topics | Tokens |
|---|---|
| Education Problems | 'college', 'online learning', 'class', 'semester', 'freshman' |
| Occupation Problems | 'lost job', 'unemployed', 'laid off', 'income', 'money', 'quit job', 'career' |
| Lonely | 'social interaction', 'interact', 'connection', 'lonely', 'friendless', 'feel alone', 'loneliness', 'friendless', 'social life', 'friendship', 'socialize', 'make friends', 'new friends', 'disconnected' |
| Fear of coronavirus | 'no mask', 'without mask', 'maskless', 'unmasked', 'grocery', 'panic', 'precautions', 'coworker', 'cough', 'exposed', 'wash', 'temperature', 'OCD' |
| Pandemic Development | 'forever', 'permanent', 'back normal', 'new normal', 'ever end', 'never ending', 'endless', 'lose hope', 'normal life' |

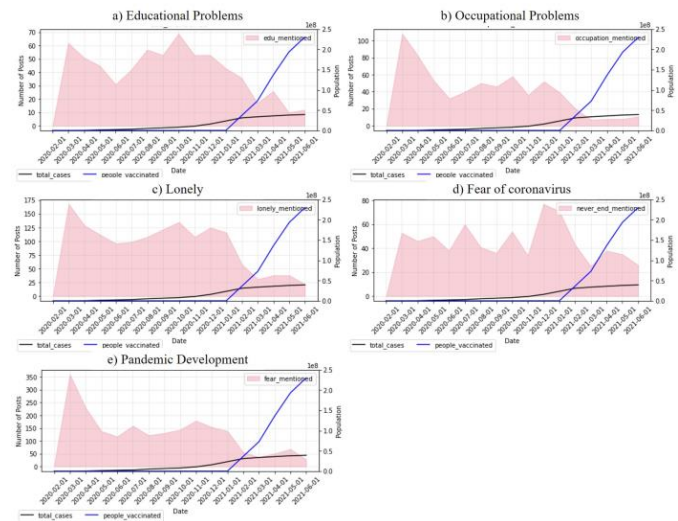

Fig. 7 Trend on number of posts mentioning each of the topics in lexical approach. a) Educational Problem, b) Occupational Problem, c) Lonely, d) Fear of coronavirus, e) Pandemic Development

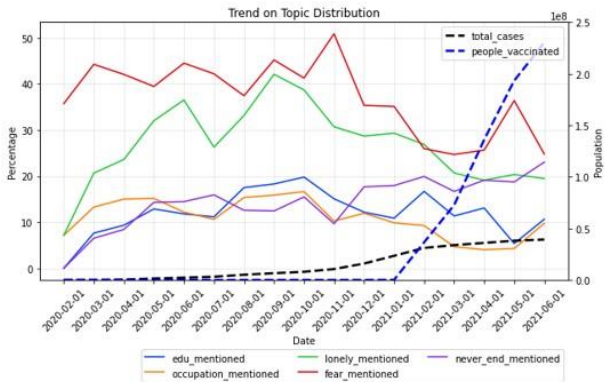

Fig. 8 This figure shows the percentage of posts mentioning each of the topics at each month. The figure illustrated which was the dominant topic at each stage of the pandemic.

## IV. DISCUSSIONS

### A. Principal Findings

In this study, with the use of topic model on Reddit data on subreddit \rCOVID19_support, we identified six topics related to the pandemic, which were "fear of coronavirus", "educational and occupational problems", "family problems", "problems related to social environment", "mental health symptoms" and "Uncertainty on development of pandemic".

The number of posts mentioning the fear of coronavirus was the highest in Mar 2020 and then dropped in Apr 2020 and May 2020, and then remained stable from May 2020 to Nov 2020. This finding is consistent with another study. Yarrington et al. [2] have studied the changes in sentiments such as anxiety, tiredness and depression during different stages of COVID-19 pandemic, with the use of data collected by a mental health app. Results in the study shows that anxiety reduced and then remain stable during the acute (12 Mar 2020 to 15 Apr 2020) and sustained (16 Apr 2020 to 06 Jul 2020) stage of stay-at-home orders.

According to the result in our study, the number of stressors related posts has reduced significantly after the vaccine started being distributed. This suggests that the mental health status was alleviated when more people were vaccinated. This could be explained by two reasons: reducing perceived risks of COVID19, and feeling hope for back to normal life. Regarding perceived risks, vaccinated people may understand vaccine could reduce the risk of infection and severity of symptoms of coronavirus, thus reduced the fear of coronavirus after being vaccinated. As vaccinated population increased, number of cases significantly dropped. More people observed the effectiveness of vaccines, and the rise on proportion of vaccinated population may have helped reduce the perceived risks of getting infected. Regarding hopes of getting back to normal life, cities announced or started the re-opening of school and economy. People may have felt less depressed about the current impacts of lockdown as they anticipated vaccines can end the pandemic.

However, the number of posts about "uncertainty about pandemic development" did not have notable drop while vaccinated population was growing. One possible cause for this phenomenon is the feeling of uncertainty was proportional to time. The idea of "pandemic is never ending" may have appeared more frequently when the pandemic lasted long. Another possible cause is people were doubtful about whether vaccine could be the solution for COVID-19. New variants of COVID-19 were continuously found and there was on-going research about the effectiveness of existing vaccines for each variant. The new variants may have intensified the worries about whether the long-term pandemic could be ended by existing vaccines. According to the result, the number of posts mentioning the thoughts of "pandemic was never ending" had a sudden increase in Apr 2021. In Apr 2021, the number of delta variant cases in U.S. was rising. This suggests the variants may cause the feeling of uncertainty and the worries of lockdown to happen again.

### B. Limitation

In this study, the prevalence of stressors was only compared with the number of cases and vaccinated population, and not compared with specific safety measures in specific cities. This was due to the anonymity on Reddit. The demographic information of users was unknown, and the dataset obtained in this study included posts that were written by users at different countries. Every city has implemented social distancing measures and lockdowns at different time, depending on the number of cases and situations of hospitalization. Due to this situation, we could not analyze the relationship of mental health status and safety measures. However, the stressor lexicon created in this study could be utilized on labelling posts in other social media platforms, such as Twitters.

## V. CONCLUSIONS

In this study, we have applied topic modelling to a dataset that contains Reddit posts in \rCOVID19_support, to identify the COVID19 related psychosocial stressors and to visualize the trend on prevalence of the stressors. Compared to the existing research which has utilized NLP techniques on social media data to study the mental health impacts of pandemic, our study focuses on stressors instead of the mental health status. Regarding the choice of dataset, the dataset used in this study include posts that were created in a time range of more than a year of pandemic. This allows us to compare the difference on prevalence of stressors before and after the vaccine has been distributed. The proposed topic model also allows monitoring the dominant stressors, which enables mental health support providers to notice the changes of stressors at different stages of the pandemic. This study has demonstrated the potential of using topic modelling on social media discussions to identify event-specific stressors, and to create dashboard to analyze and monitor the trend. We hope the findings in this study could provide insights for healthcare providers and social workers to handle the needs of Covid-19 related mental health support. Furthermore, we hope the NLP techniques used in this study may be applied for analyzing the psychosocial stressors and create corresponding lexicons of future events such as pandemic, protests or financial crisis.